\documentclass[letterpaper]{article} 
\usepackage{aaai23-r2hcai}  
\usepackage{times}  
\usepackage{helvet}  
\usepackage{courier}  
\usepackage[hyphens]{url}  
\usepackage{graphicx} 
\def\UrlFont{\rm}  
\usepackage{natbib}  
\usepackage{caption} 
\usepackage{xcolor}
\usepackage{algorithm}
\usepackage{algorithmic}
\usepackage{subcaption}
\usepackage{graphicx}
\usepackage{mathtools}
\usepackage{amsmath,amssymb}
\usepackage{breqn}
\usepackage{physics}

\DeclareMathOperator{\Rh}{\mathcal{\tilde{R}}^h}
\DeclareMathOperator{\G}{\mathcal{G}}
\newcommand{\Rpsi}{R^{h}_{\psi}}

\newcommand{\Rvec}{\vectorbold{R}_{\psi}^h}

\usepackage{newfloat}
\usepackage{listings}
\DeclareCaptionStyle{ruled}{labelfont=normalfont,labelsep=colon,strut=off} 
\floatstyle{ruled}
\newfloat{listing}{tb}{lst}{}
\floatname{listing}{Listing}
\usepackage{fancyhdr}
\fancypagestyle{firstpagehf}
{
   \fancyhf{}
   \fancyhead[HC]{The AAAI 2023 Workshop on Representation Learning for Responsible Human-Centric AI (R$^2$HCAI)}
   \fancyfoot[C]{\thepage}
}

\title{A State Augmentation based approach to Reinforcement Learning from Human Preferences}
\author{
   Mudit Verma\textsuperscript{\rm 1} \quad
   Subbarao Kambhampati \textsuperscript{\rm 1} \quad
}
\affiliations{
    \textsuperscript{\rm 1} SCAI, Arizona State University\\
    muditverma@asu.edu, rao@asu.edu
    
}

\usepackage{bibentry}

\begin{document}
\thispagestyle{firstpagehf}
\maketitle

\begin{abstract}
Reinforcement Learning has suffered from poor reward specification, and issues for reward hacking even in simple enough domains. Preference Based Reinforcement Learning attempts to solve the issue by utilizing binary feedbacks on queried trajectory pairs by a human in the loop indicating their preferences about the agent's behavior to learn a reward model. In this work, we present a state augmentation technique that allows the agent's reward model to be robust and follow an invariance consistency that significantly improved performance, i.e. the reward recovery and subsequent return computed using the learned policy over our baseline PEBBLE. We validate our method on three domains, Mountain Car, a locomotion task of Quadruped-Walk, and a robotic manipulation task of Sweep-Into, and find that using the proposed augmentation the agent not only benefits in the overall performance but does so, quite early in the agent's training phase.

\end{abstract}

\section{Introduction}
Machine Learning \cite{pouyanfar2018survey, verma2020fine, verma2019novel}, specifically Deep Reinforcement Learning \cite{mnih2015human, verma2019making, arulkumaran2017deep} has been quite successful for several tasks \cite{openaigym} that especially those with high dimensional states and action spaces. However, much of the credit to the RL agent's success is attributed to the reward function specification \cite{ird, expressivity, secret, verma2019making}. Among several challenges in designing a ``good reward" \cite{ird} function that allows the RL agent to learn trustworthy policies \cite{zahedi2021trust, zahedi2022modeling} like sparsity of the reward, balancing scales of reward signals, etc, one major issue that researchers have found is the inability of specifying the reward function, to begin with, \cite{reward-hacking-example}. To allow humans in the loop to specify their preferences over the agent's goals and behavior, Preference-Based Reinforcement Learning (PbRL) \cite{wilson2012} utilizes binary evaluative signals over trajectory pairs given by the human in the loop to convey their preferences to learn the human's reward model and subsequently learn a usable policy over it.

Preference Based Reinforcement Learning allows human in the loop to specify their preferences in the form of binary feedback over agent-generated trajectory pair queries. This can prove to be quite beneficial for several reasons. First, the human in the loop gets to realize how the trajectories ``look" like and does not have to imagine these trajectories thereby taking into account human's cognitive limitations. Second, the human in the loop need not be an expert engineer and is not required to be aware of the underlying agent representation to hand-design a reward function and instead can use the image as the lingua franca \cite{expand, guan2020explanation, lingua-franca} to specify their preferences. Finally, issues of reward hacking and ``cheat-behaviors" have been noted by researchers in prior works and PbRL can serve as a potential alternative to reward engineering to tackle these issues \cite{reward-hacking-example, reward-hacking-init}.

Recent works in PbRL have focused on improving the query strategy \cite{pebble, christiano}, exploration method \cite{pebble}, priors regarding the reward function \cite{prior}, semi-supervised learning methods \cite{surf} and temporal data augmentation \cite{surf}. However, to the best of our knowledge, none of the existing methods have attempted to provide a tailored solution for the problem PbRL with image-based state representations. In this work, we specifically focus on the problem of learning from human preferences via binary feedback on trajectory pair queries with a pixel-based image state representation of the agent \cite{mnih2015human, mnih2013playing}. This is particularly helpful as in most of the recent PbRL works the lingua franca used to communicate with the human in the loop are image-based representations. Motivated by prior works in explainable AI \cite{adv-conf} and Advisable reinforcement learning \cite{blackbox, expand, guan2020explanation, verma2021perfect}, we present a data augmentation technique over image-based state representation of the agent that shows significant improvements in the agent's reward recovery and subsequently high return collected by the agent's learned policy over the learned reward model for three continuous control tasks, Mountain-Car, a robotic manipulation task Sweep-Into, and a locomotion task Quadruped-Walk.

Data augmentation has been explored for various reasons in the context of machine learning and artificial intelligence \cite{aug_survey, data-aug, expand}, of which the foremost reasons are the robustness of the learned model \cite{cobbe2019quantifying} and to impose consistencies \cite{consistency-3, consistency-1, consistency-2}. In addition to the requirements of the robustness of the reward model and to impose certain invariance consistencies we leverage the observation that a good reward model should ``focus" on regions of importance and this should be reflected by the model's representation space. We combine the requirements of robustness, invariance consistency and to have the reward model ``focus" on regions of importance into a single data augmentation technique and showcase the benefits across three continuous control tasks (OpenAI Gym's Mountain Car \cite{openaigym}, DM Control's locomotion task of Quadruped-Walk \cite{dmcontrol} and Metaworld's robotic manipulation task of Sweep-Into \cite{metaworld}) and show that using such an augmentation technique can help with an early boost in learning performance at zero cost to the human in the loop.

\section{Background}
We have an agent $M$ that interacts with the environment $\mathcal{E}$ by taking an action $a \sim \mathcal{A}$ on a state $s$. Typical reinforcement learning frameworks assume an underlying Markov Decision Process (MDP) as the tuple $<\mathcal{S, T, A, R}>$ where $\mathcal{S}$ is the state space, $\mathcal{T}$ is the transition function, $\mathcal{A}$ is the set of permissible actions and $\mathcal{R}$ is the task reward function, however Preference-based Reinforcement learning updates this tuple by replacing $\mathcal{R}$ with $\Rh$, the human reward model, as the tuple $<\mathcal{S, T, A,} \Rh>$. 

The goal of PbRL is to approximate the human reward model $\Rh$ with a parameterized function approximator $\Rpsi$. The agent, $M$, queries the human in the loop with a trajectory pair $\tau_0, \tau_1$, $\tau_i = \{(s_k, a_k), (s_{k+1}, a_{k+1} \cdots (s_{k+H}, a_{k+H}))\}$ and receives a binary feedback $y \in \{0,1\}$ indicating the human's preferred trajectory, i.e. $y=0$ if $\tau_0$ is preferred over $\tau_1$ and vice versa. Such feedbacks along with the queried trajectories are stored in a dataset $D_\tau$ as tuples $(\tau_0, \tau_1, y)$. Recent PbRL works have leveraged the Bradley Terry model \cite{bradley-terry} to compute the probability of one trajectory being preferred over another. With means to computing this probability, PbRL methods treat essentially solve the reward learning problem via a classification problem where the trajectory with a higher approximated sum of rewards (or the return) is predicted to be the human preferred trajectory and a reward assignment to the constituent states of $\tau_0, \tau_1$ that achieves high accuracy in this supervised learning task is taken to be the approximate human reward model. As typical in binary classification problems of supervised learning, this is done by minimizing the cross-entropy between the predictions and ground truth human labels as follows:

\begin{equation} \label{eq1}
\begin{split}
    \mathcal{L}_{CE}=  \displaystyle \mathop{- \mathbb{E}}_{(\tau_0, \tau_1, y)\sim \mathcal{D}} 
    & [y(0)\text{log}P_\psi[\tau_0 \succ \tau_1] \\
    & + y(1)\text{log}P_\psi[\tau_1 \succ \tau_0]]
\end{split}
\end{equation}

where probabilities $P_\psi$ are computed using the approximated reward function $\Rpsi$ as : 
 \begin{equation}
     P_{\psi}[\tau_0 \succ \tau_1] = \frac{\exp(\sum_{t}\Rpsi(s_t^0, a_t^0))}{\sum_{i \in \{0,1\}} \exp(\sum_{t}\Rpsi(s_t^i, a_t^i))}
 \end{equation}

\section{Method}
The key contribution of this work is to allow the reward model to better utilize the queried trajectory pairs in the context of image-based state representation of the RL agent $M$. We aim to leverage three key benefits from a single state augmentation technique, namely, \textbf{robustness}: a key indicator for generalization of the learned rewards which becomes all the more necessary with the aim of reducing the human feedback sample complexity, \textbf{invariance}: to allow the agent to successfully learn the state representation in the reward model such that it is invariant to perturbations to regions in the image space which is not important, and finally, \textbf{motion-based-importance}: that marks regions in the image observations of the agent where a change has occurred intending to motivate the agent to ``focus" on such potential regions while predicting the reward.

As noted by several prior works in the area of explainable AI and advisable RL, image observations are expressed as the pair $<I_C, I_S>$ where $I_C$ are the pixels denoting the content of the image and $I_S$ denotes the pixels that inform about the style of the image. Typically, pixels related to the style of the image do not offer any additional information than just $I_C$ and $I_C$ would conceptually be similar to the effective dimensionality of the image observation required for the task in concern. Prior literature has utilized either human annotations to assume $I_C$ or attempted to use explanatory techniques to inform users about $I_C$. In this work, we use the ``motion-based-importance" insight to capture $I_C$. For an image observation and agent state $I$, let's say $I_1, I_2 \cdots I_n$ are observations such that there exists some action $a_i$ which when taken on $I_j \; j \in {1,2 \cdots n}$ can transition to $I$. Then the image mask created by the union (of boolean mask matrices) of all the differences between $I_j$ and $I$ will contribute to $I_C$, i.e.
\begin{equation}
\label{eq:mask}
    \mathbb{M}(I, \{I_1, I_2 \cdots I_n\}) = \displaystyle \bigcup_{j \in \{1,2 \cdots n\}} \mathbb{M}(I, I_j)
\end{equation}

\begin{equation}
\label{eq:mask_single}
    \mathbb{M}(I, I_j) = 
    \begin{cases}
    & 1, I(x,y) \neq I_j(x,y)\\ 
    & 0, \text{otherwise}\\
    \end{cases}
\end{equation}

As the access to all predecessor states $I_j$ for every $I$ may entail high exploration and storage requirements, it can be approximated by using one predecessor observation $I_{t-1}$ at a time to create the mask $\mathbb{M}$ for observation $I_{t}$ in a trajectory $\tau = <I_0, I_1, I_2 \cdots I_n>$.

With equation \ref{eq:mask} giving us the mask (as a proxy for all the pixels referring to the content $I_C$ of an image $I$), we can utilize it to inject our requirements of invariance. A popular means of doing so is via perturbations of the style pixels. We follow, \cite{expand, greydanus} which argues for using Gaussian perturbations as would still preserve the texture of the remaining images, however, domain-specific perturbations such as a change in objects in the background are also applicable but require extra information. Gaussian perturbations using the mask in  equation \ref{eq:mask_single} can be defined as : 
\begin{equation}
\label{eq:gaussian}
\begin{split}
    \phi(I, I_j, \mathbb{M}) = \;
    & I \odot (1 - \mathbb{M}(I, I_j) \\
    & + \G(I, \sigma_{\G}) \odot \mathbb{M}(I, I_j) \\ 
\end{split}
\end{equation}

In the equation \ref{eq:gaussian}, $\G$ refers to the convolution of a gaussian filter over the image $I$ with standard deviation $\sigma_{\G}$. $\odot$ is the Hadamard product operation between the input image or the gaussian blurred image and the binary mask matrix $\mathbb{M}$. Finally, from the dataset of preference feedback $D_{\tau} = <\tau_0^i, \tau_1^i, y^i>$, we can pick each trajectory $\tau_0^i$ or $\tau_1^i$ and produce augmented trajectories by augmenting individual states and chaining those states together to form the new augmented trajectory. We augment a trajectory $\tau = <s_0, s_1, ... s_n>$ as, 
\begin{equation}
    \tau_g = <s_0, \phi_{\mathbb{M}}(s_1, s_0), \phi_{ \mathbb{M}}(s_2, s_1) \cdots \phi_{\mathbb{M}}(s_n, s_{n-1})> 
\end{equation}

and produce multiple augmented trajectories by varying the standard deviation $\sigma_{\G}$ of the gaussian filter. Hence for a set of standard deviations $\G = \{\sigma_1, \sigma_2 \cdots \sigma_{k-1}\}$ we can generate $k-1$ augmented trajectories from the original trajectory $\tau$ as, $\{\tau_g^1, \tau_g^2 \cdots \tau_g^{k-1}\}$ to get a total of $k$ trajectories (one original $\tau$, and $k-1$ augmented $\tau_g^i$).

We impose our invariance consistency requirement via the following loss over the reward model $\Rpsi$, 

\begin{equation}
    \mathcal{L}_{I} (\tau) = \frac{1}{|\G|}\sum_{i \sim \G} \norm{\Rvec(\tau) - \Rvec(\tau_g^i)}_{2}
\end{equation}
where $\Rvec(\tau) = \begin{bmatrix} \Rpsi(s_0) & \Rpsi(s_1) & \cdots & \Rpsi(s_{T}) \end{bmatrix}^T$ is a vector of predicted rewards over the states in the trajectory. This reward $\mathcal{L}_{I}$ helps with both robustness and invariance consistency as shown in section \ref{sec:results}. This final reward model update loss is a linear combination of this invariance loss and the cross entropy loss as,
\begin{equation}
    \mathcal{L}_{reward} = \lambda_{CE}\mathcal{L}_{CE} + \lambda_{I}\mathcal{L}_{I}
\end{equation}

Finally, we also treat augmented trajectories as additional data points for training the cross entropy loss $\mathcal{L}_{CE}$. Since we optimize using stochastic gradient descent, we first sample a batch of trajectory pairs and labels from $D_\tau$ and then use augmented versions of these trajectories as additional data points within this batch.

\section{Experiments}
\label{sec:results}

\begin{figure}[h]
    \centering
    \includegraphics[width=0.45\textwidth]{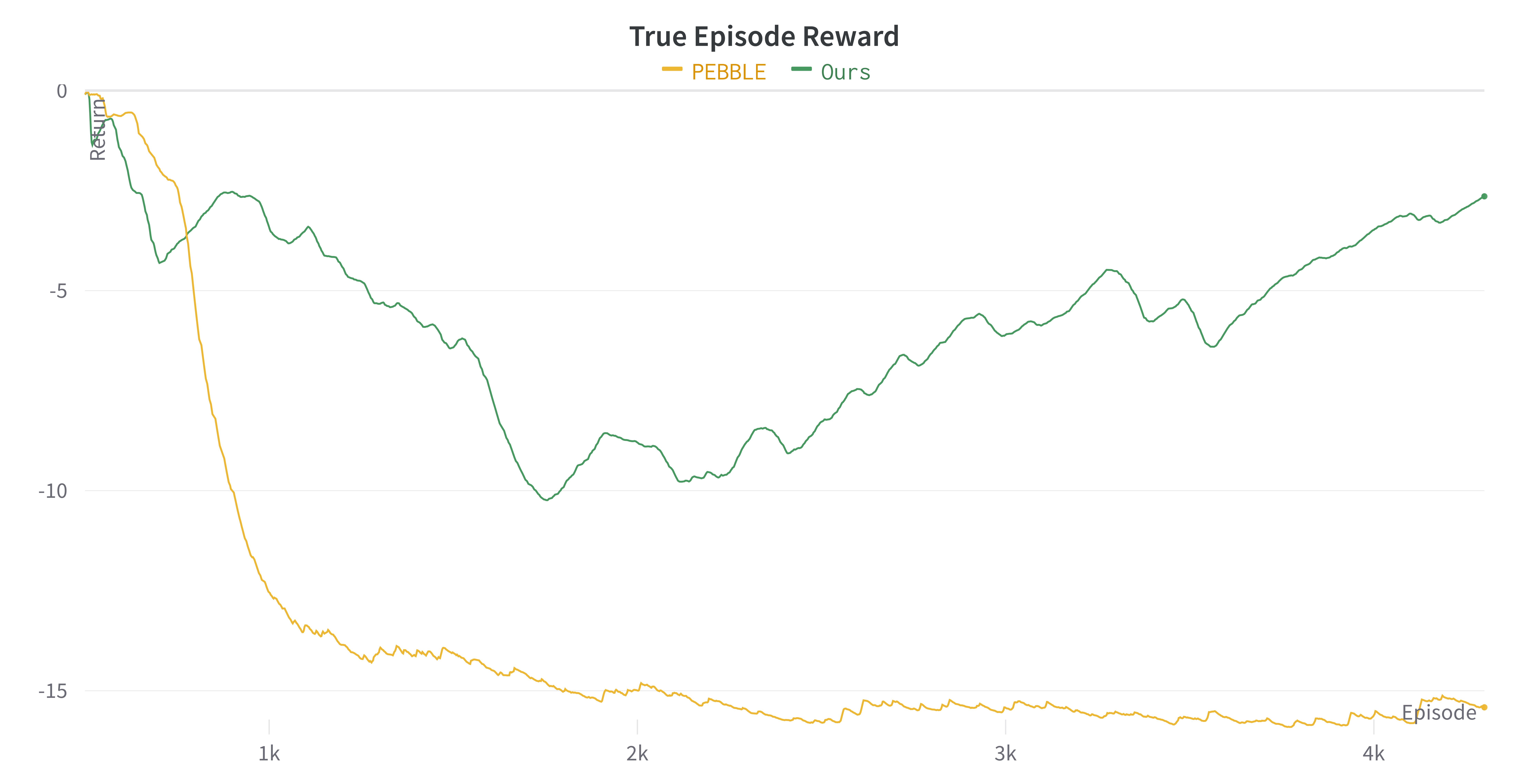}
    \caption{Evaluation curves on Mountain Car Continuous Control as measured on the ground truth human reward $\Rh$.}
    \label{fig:mc}
\end{figure}

\begin{figure}[h]
    \centering
    \includegraphics[width=0.45\textwidth]{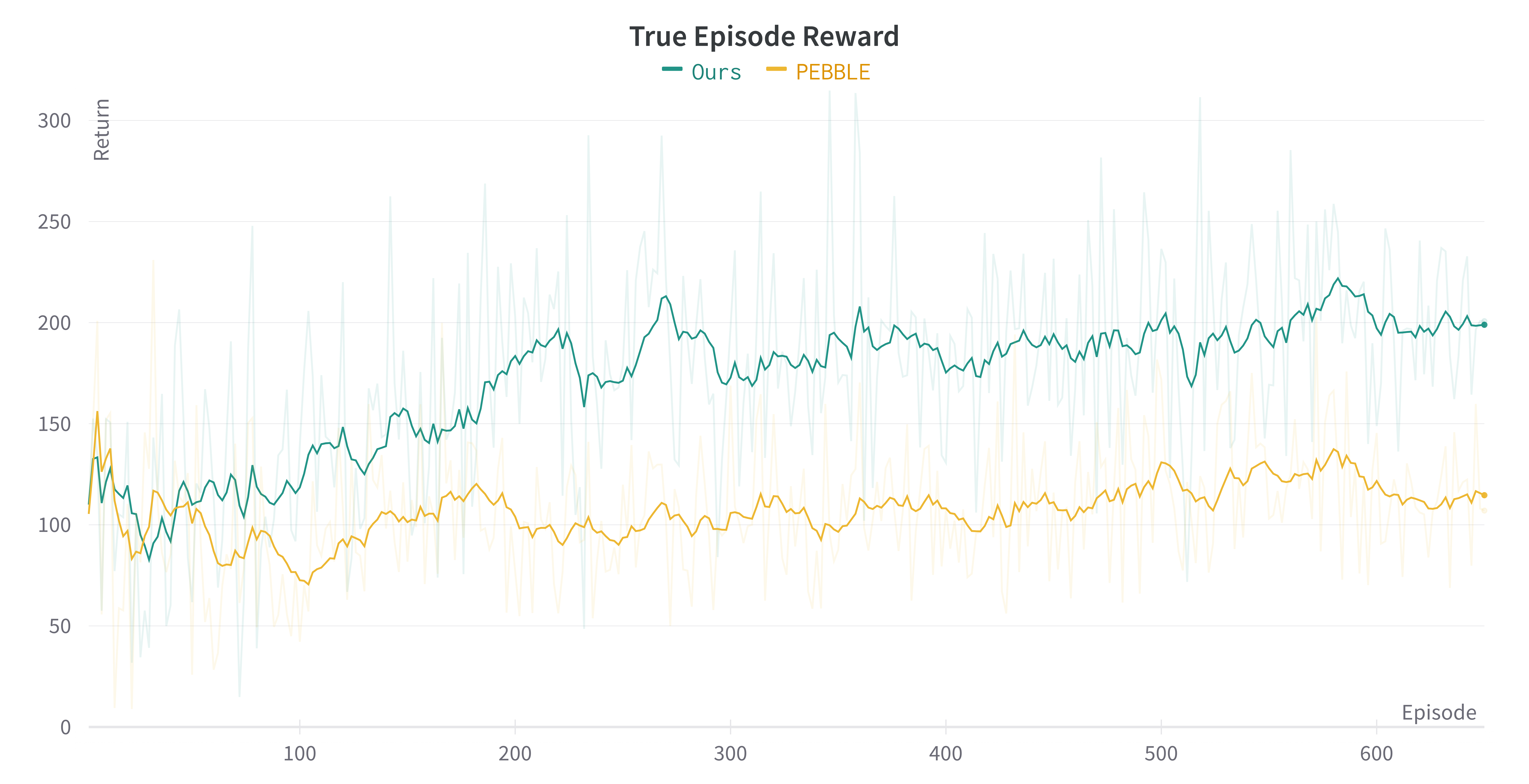}
    \caption{Evaluation curves on the locomotion task of Quadruped as measured on the ground truth human reward $\Rh$.}
    \label{fig:quad}
\end{figure}

\begin{figure}[h]
    \centering
    \includegraphics[width=0.45\textwidth]{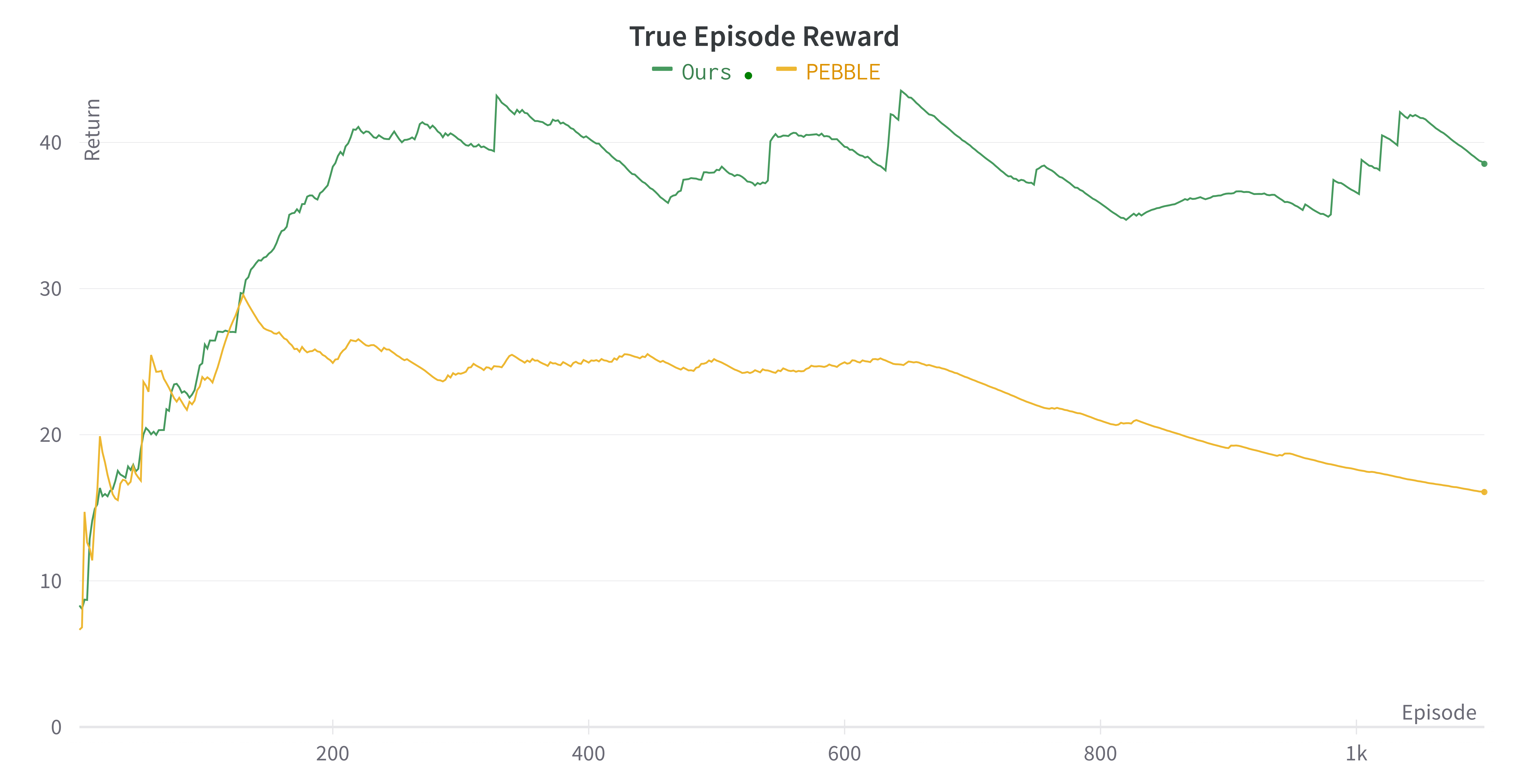}
    \caption{Evaluation curves on the robotic manipulation task of Sweep Into as measured on the ground truth human reward $\Rh$.}
    \label{fig:sweep}
\end{figure}

In the preliminary investigation, we were interested to answer whether the proposed augmentation technique can outperform the current state-of-the-art PbRL method (PEBBLE).

We validate our results on three continuous control domains on pixel-based state representations (with standard preprocessing \cite{mnih2015human, mnih2013playing}), i.e. Mountain Car (\texttt{MountainCarContinuous-v0}) by OpenAI gym \cite{openaigym}, one locomotion task of Quadruped Walk \texttt{quadruped-walk} from MuJoCo \cite{dmcontrol} and one robotic manipulation task Sweep Into \texttt{sweep-into-v2} from Metaworld \cite{metaworld}. We use PEBBLE as the backbone Preference-based Reinforcement Learning algorithm and update the $\mathcal{L}_{reward} = \lambda_{CE}\mathcal{L}_{CE} + \lambda_{I}\mathcal{L}_{I}$ with $\lambda_{CE} = 1, \lambda_{I} = 0.6$, and as mentioned before, use the augmented trajectories data for training the cross-entropy loss as well by appending to the dataset $D_\tau$ the following tuples $<\tau_g^0, \tau_g^1, y>_{i=1}^{i=|\G|}$ for each $<\tau_0, \tau_1, y>$. PEBBLE uses SAC as the Reinforcement Learning algorithm underneath, and we used the same hyperparameters used for Sweep Into and Quadruped-Walk as suggested in PEBBLE (although they primarily showcase results on low-level states instead of image observations). We gave 1000 feedbacks for Quadruped-Walk and Mountain Car, and 10000 feedbacks for Sweep-Into. SURF shows results on pixel-based inputs for PEBBLE achieve similar performance as our implementation of PEBBLE baseline. Finally, as suggested in prior works, we use an oracular approach to evaluation where the rewards that come packaged with these environments are assumed to be human's reward model, and the extent to which a PbRL method can recover this underlying reward model is used as the success criterion.

\subsection{Results}
In order to realize how well a PbRL framework learned the reward model, we evaluate the policy $\pi$ learned on $\Rpsi$ over the human's underlying reward model $\Rh$. Any improvements over the performance of the learned policy on the underlying human reward model $\Rh$ would indicate a more meaningful learned reward model $\Rpsi$.

\textbf{Mountain Car}: is used as a toy domain to verify our claims. The domain consists of a car placed at the bottom of a sinusoidal valley with the goal to strategically accelerate the car (action space is acceleration between -1 to 1) to reach the top of the right hill \cite{openaigym}. Figure \ref{fig:mc}. We find that with very few feedbacks taken in initial epochs, our approach can learn a reasonable reward model and with more human feedbacks it outperforms the baseline.
 
\textbf{Quadruped-Walk}: is a quadruped (two bipedals) robot, with four legs each having three actuators (a total of 12 continuous actions) with the goal of walking on a flat surface. Figure \ref{fig:quad} shows that our approach outperforms PEBBLE baseline on this locomotion task and as seen before, the highest performance boost occurs within the initial training episodes.

\textbf{Sweep-Into}: is a robotic arm in front of a puck placed on a table with the goal to ``sweep" the puck into the goal location. The puck positions are randomized at the start of each episode (as per the default package setup). We find similar results as in our other two domains, where the proposed state/trajectory augmentation (and subsequently the updated reward loss) outperforms the baseline PEBBLE. Additionally, we notice that the augmentations provide a significant boost to the performance early in the training process and help the agent maintain this performance gain throughout.

\section{Discussion}
In this work, we presented a state augmentation technique tailored for image-based Preference-based Reinforcement Learning. The proposed augmentation utilizes the insight that regions of the image observation that update upon a transition (when an action is taken), are at least a subset of all the ``content" available in the image observation. This allows the reward model being learned to be cognizant of potential regions of the image observation that are likely to be updated in future steps (and history) while predicting the rewards. Augmented trajectories are treated as additional query data points to train the cross entropy loss along with the proposed invariance loss that maintains prediction consistency \cite{expand, consistency-1}. The benefits of the approach were validated on three continuous control domains, OpenAI gym's Mountain Car, a locomotion domain: DM Control's Quadruped-Walk, and finally a robotic manipulation domain: Meta World's Sweep-Into. 

Future work includes an exhaustive evaluation of the proposed augmentation across more domains. We also plan to evaluate the benefits of the proposed work when used with other PbRL techniques and paradigms. Finally, we would also like to study the advantages of using domain-dependent perturbations.

\section*{Acknowledgements}
Kambhampati's research is supported by the J.P. Morgan Faculty Research Award, ONR grants N00014-16-1-2892, N00014-18-1-2442, N00014-18-1-2840, N00014-9-1-2119, AFOSR grant FA9550-18-1-0067 and DARPA SAIL-ON grant W911NF19-2-0006. 

\bibliography{aaai23}

\begin{thebibliography}{36}
\providecommand{\natexlab}[1]{#1}

\bibitem[{Abel et~al.(2021)Abel, Dabney, Harutyunyan, Ho, Littman, Precup, and
  Singh}]{expressivity}
Abel, D.; Dabney, W.; Harutyunyan, A.; Ho, M.~K.; Littman, M.; Precup, D.; and
  Singh, S. 2021.
\newblock On the expressivity of markov reward.
\newblock \emph{Advances in Neural Information Processing Systems}, 34:
  7799--7812.

\bibitem[{Arulkumaran et~al.(2017)Arulkumaran, Deisenroth, Brundage, and
  Bharath}]{arulkumaran2017deep}
Arulkumaran, K.; Deisenroth, M.~P.; Brundage, M.; and Bharath, A.~A. 2017.
\newblock Deep reinforcement learning: A brief survey.
\newblock \emph{IEEE Signal Processing Magazine}, 34(6): 26--38.

\bibitem[{Berthelot et~al.(2019)Berthelot, Carlini, Cubuk, Kurakin, Sohn,
  Zhang, and Raffel}]{consistency-2}
Berthelot, D.; Carlini, N.; Cubuk, E.~D.; Kurakin, A.; Sohn, K.; Zhang, H.; and
  Raffel, C. 2019.
\newblock Remixmatch: Semi-supervised learning with distribution alignment and
  augmentation anchoring.
\newblock \emph{arXiv preprint arXiv:1911.09785}.

\bibitem[{Bradley and Terry(1952)}]{bradley-terry}
Bradley, R.~A.; and Terry, M.~E. 1952.
\newblock Rank analysis of incomplete block designs: I. The method of paired
  comparisons.
\newblock \emph{Biometrika}, 39(3/4): 324--345.

\bibitem[{Brockman et~al.(2016)Brockman, Cheung, Pettersson, Schneider,
  Schulman, Tang, and Zaremba}]{openaigym}
Brockman, G.; Cheung, V.; Pettersson, L.; Schneider, J.; Schulman, J.; Tang,
  J.; and Zaremba, W. 2016.
\newblock Openai gym.
\newblock \emph{arXiv preprint arXiv:1606.01540}.

\bibitem[{Christiano et~al.(2017)Christiano, Leike, Brown, Martic, Legg, and
  Amodei}]{christiano}
Christiano, P.~F.; Leike, J.; Brown, T.; Martic, M.; Legg, S.; and Amodei, D.
  2017.
\newblock Deep reinforcement learning from human preferences.
\newblock \emph{Advances in neural information processing systems}, 30.

\bibitem[{Cobbe et~al.(2019)Cobbe, Klimov, Hesse, Kim, and
  Schulman}]{cobbe2019quantifying}
Cobbe, K.; Klimov, O.; Hesse, C.; Kim, T.; and Schulman, J. 2019.
\newblock Quantifying generalization in reinforcement learning.
\newblock In \emph{International Conference on Machine Learning}, 1282--1289.
  PMLR.

\bibitem[{Ferret et~al.(2020)Ferret, Marinier, Geist, and Pietquin}]{secret}
Ferret, J.; Marinier, R.; Geist, M.; and Pietquin, O. 2020.
\newblock Self-Attentive Credit Assignment for Transfer in Reinforcement
  Learning.

\bibitem[{Greydanus et~al.(2018)Greydanus, Koul, Dodge, and Fern}]{greydanus}
Greydanus, S.; Koul, A.; Dodge, J.; and Fern, A. 2018.
\newblock Visualizing and understanding atari agents.
\newblock In \emph{International conference on machine learning}, 1792--1801.
  PMLR.

\bibitem[{Guan et~al.(2021)Guan, Verma, Guo, Zhang, and Kambhampati}]{expand}
Guan, L.; Verma, M.; Guo, S.~S.; Zhang, R.; and Kambhampati, S. 2021.
\newblock Widening the pipeline in human-guided reinforcement learning with
  explanation and context-aware data augmentation.
\newblock \emph{Advances in Neural Information Processing Systems}, 34:
  21885--21897.

\bibitem[{Guan, Verma, and Kambhampati(2020)}]{guan2020explanation}
Guan, L.; Verma, M.; and Kambhampati, S. 2020.
\newblock Explanation augmented feedback in human-in-the-loop reinforcement
  learning.
\newblock \emph{arXiv preprint arXiv:2006.14804}.

\bibitem[{Hadfield-Menell et~al.(2017)Hadfield-Menell, Milli, Abbeel, Russell,
  and Dragan}]{ird}
Hadfield-Menell, D.; Milli, S.; Abbeel, P.; Russell, S.~J.; and Dragan, A.
  2017.
\newblock Inverse reward design.
\newblock \emph{Advances in neural information processing systems}, 30.

\bibitem[{Kambhampati et~al.(2022)Kambhampati, Sreedharan, Verma, Zha, and
  Guan}]{lingua-franca}
Kambhampati, S.; Sreedharan, S.; Verma, M.; Zha, Y.; and Guan, L. 2022.
\newblock Symbols as a lingua franca for bridging human-ai chasm for
  explainable and advisable ai systems.
\newblock In \emph{Proceedings of the AAAI Conference on Artificial
  Intelligence}, volume~36, 12262--12267.

\bibitem[{Krakovna et~al.(2020)Krakovna, Uesato, Mikulik
  et~al.}]{reward-hacking-init}
Krakovna, V.; Uesato, J.; Mikulik, V.; et~al. 2020.
\newblock Specification gaming: The flip side of AI ingenuity| DeepMind.

\bibitem[{Lee, Smith, and Abbeel(2021)}]{pebble}
Lee, K.; Smith, L.; and Abbeel, P. 2021.
\newblock Pebble: Feedback-efficient interactive reinforcement learning via
  relabeling experience and unsupervised pre-training.
\newblock \emph{arXiv preprint arXiv:2106.05091}.

\bibitem[{Mnih et~al.(2013)Mnih, Kavukcuoglu, Silver, Graves, Antonoglou,
  Wierstra, and Riedmiller}]{mnih2013playing}
Mnih, V.; Kavukcuoglu, K.; Silver, D.; Graves, A.; Antonoglou, I.; Wierstra,
  D.; and Riedmiller, M. 2013.
\newblock Playing atari with deep reinforcement learning.
\newblock \emph{arXiv preprint arXiv:1312.5602}.

\bibitem[{Mnih et~al.(2015)Mnih, Kavukcuoglu, Silver, Rusu, Veness, Bellemare,
  Graves, Riedmiller, Fidjeland, Ostrovski et~al.}]{mnih2015human}
Mnih, V.; Kavukcuoglu, K.; Silver, D.; Rusu, A.~A.; Veness, J.; Bellemare,
  M.~G.; Graves, A.; Riedmiller, M.; Fidjeland, A.~K.; Ostrovski, G.; et~al.
  2015.
\newblock Human-level control through deep reinforcement learning.
\newblock \emph{nature}, 518(7540): 529--533.

\bibitem[{Park et~al.(2022)Park, Seo, Shin, Lee, Abbeel, and Lee}]{surf}
Park, J.; Seo, Y.; Shin, J.; Lee, H.; Abbeel, P.; and Lee, K. 2022.
\newblock SURF: Semi-supervised Reward Learning with Data Augmentation for
  Feedback-efficient Preference-based Reinforcement Learning.
\newblock \emph{arXiv preprint arXiv:2203.10050}.

\bibitem[{Pouyanfar et~al.(2018)Pouyanfar, Sadiq, Yan, Tian, Tao, Reyes, Shyu,
  Chen, and Iyengar}]{pouyanfar2018survey}
Pouyanfar, S.; Sadiq, S.; Yan, Y.; Tian, H.; Tao, Y.; Reyes, M.~P.; Shyu,
  M.-L.; Chen, S.-C.; and Iyengar, S.~S. 2018.
\newblock A survey on deep learning: Algorithms, techniques, and applications.
\newblock \emph{ACM Computing Surveys (CSUR)}, 51(5): 1--36.

\bibitem[{Shorten and Khoshgoftaar(2019)}]{aug_survey}
Shorten, C.; and Khoshgoftaar, T.~M. 2019.
\newblock A survey on image data augmentation for deep learning.
\newblock \emph{Journal of big data}, 6(1): 1--48.

\bibitem[{Sohn et~al.(2020)Sohn, Berthelot, Carlini, Zhang, Zhang, Raffel,
  Cubuk, Kurakin, and Li}]{consistency-3}
Sohn, K.; Berthelot, D.; Carlini, N.; Zhang, Z.; Zhang, H.; Raffel, C.~A.;
  Cubuk, E.~D.; Kurakin, A.; and Li, C.-L. 2020.
\newblock Fixmatch: Simplifying semi-supervised learning with consistency and
  confidence.
\newblock \emph{Advances in neural information processing systems}, 33:
  596--608.

\bibitem[{Sreedharan et~al.(2020)Sreedharan, Soni, Verma, Srivastava, and
  Kambhampati}]{blackbox}
Sreedharan, S.; Soni, U.; Verma, M.; Srivastava, S.; and Kambhampati, S. 2020.
\newblock Bridging the Gap: Providing Post-Hoc Symbolic Explanations for
  Sequential Decision-Making Problems with Inscrutable Representations.
\newblock \emph{arXiv preprint arXiv:2002.01080}.

\bibitem[{Srinivas, Laskin, and Abbeel(2020)}]{data-aug}
Srinivas, A.; Laskin, M.; and Abbeel, P. 2020.
\newblock Curl: Contrastive unsupervised representations for reinforcement
  learning.
\newblock \emph{arXiv preprint arXiv:2004.04136}.

\bibitem[{Tunyasuvunakool et~al.(2020)Tunyasuvunakool, Muldal, Doron, Liu,
  Bohez, Merel, Erez, Lillicrap, Heess, and Tassa}]{dmcontrol}
Tunyasuvunakool, S.; Muldal, A.; Doron, Y.; Liu, S.; Bohez, S.; Merel, J.;
  Erez, T.; Lillicrap, T.; Heess, N.; and Tassa, Y. 2020.
\newblock dm\_control: Software and tasks for continuous control.
\newblock \emph{Software Impacts}, 6: 100022.

\bibitem[{Vamplew et~al.(2018)Vamplew, Dazeley, Foale, Firmin, and
  Mummery}]{reward-hacking-example}
Vamplew, P.; Dazeley, R.; Foale, C.; Firmin, S.; and Mummery, J. 2018.
\newblock Human-aligned artificial intelligence is a multiobjective problem.
\newblock \emph{Ethics and Information Technology}, 20(1): 27--40.

\bibitem[{Verma et~al.(2019{\natexlab{a}})Verma, Bhambri, Gupta, and
  Buduru}]{verma2019making}
Verma, M.; Bhambri, S.; Gupta, S.; and Buduru, A.~B. 2019{\natexlab{a}}.
\newblock Making Smart Homes Smarter: Optimizing Energy Consumption with Human
  in the Loop.
\newblock \emph{arXiv preprint arXiv:1912.03298}.

\bibitem[{Verma and Buduru(2020)}]{verma2020fine}
Verma, M.; and Buduru, A.~B. 2020.
\newblock Fine-grained language identification with multilingual CapsNet model.
\newblock In \emph{2020 IEEE Sixth International Conference on Multimedia Big
  Data (BigMM)}, 94--102. IEEE.

\bibitem[{Verma, Kharkwal, and Kambhampati(2022)}]{adv-conf}
Verma, M.; Kharkwal, A.; and Kambhampati, S. 2022.
\newblock Advice Conformance Verification by Reinforcement Learning agents for
  Human-in-the-Loop.
\newblock \emph{arXiv preprint arXiv:2210.03455}.

\bibitem[{Verma and Metcalf(2022)}]{prior}
Verma, M.; and Metcalf, K. 2022.
\newblock Symbol Guided Hindsight Priors for Reward Learning from Human
  Preferences.
\newblock \emph{arXiv preprint arXiv:2210.09151}.

\bibitem[{Verma et~al.(2021)Verma, Shah, Nayyar, and Hanni}]{verma2021perfect}
Verma, M.; Shah, N.; Nayyar, R.~K.; and Hanni, A. 2021.
\newblock Perfect Observability is a Myth: Restraining Bolts in the Real World.

\bibitem[{Verma et~al.(2019{\natexlab{b}})Verma, Sinha, Goyal, Verma, and
  Susan}]{verma2019novel}
Verma, M.; Sinha, P.; Goyal, K.; Verma, A.; and Susan, S. 2019{\natexlab{b}}.
\newblock A novel framework for neural architecture search in the hill climbing
  domain.
\newblock In \emph{2019 IEEE Second International Conference on Artificial
  Intelligence and Knowledge Engineering (AIKE)}, 1--8. IEEE.

\bibitem[{Wilson, Fern, and Tadepalli(2012)}]{wilson2012}
Wilson, A.; Fern, A.; and Tadepalli, P. 2012.
\newblock A bayesian approach for policy learning from trajectory preference
  queries.
\newblock \emph{Advances in neural information processing systems}, 25.

\bibitem[{Xie et~al.(2020)Xie, Dai, Hovy, Luong, and Le}]{consistency-1}
Xie, Q.; Dai, Z.; Hovy, E.; Luong, T.; and Le, Q. 2020.
\newblock Unsupervised data augmentation for consistency training.
\newblock \emph{Advances in Neural Information Processing Systems}, 33:
  6256--6268.

\bibitem[{Yu et~al.(2020)Yu, Quillen, He, Julian, Hausman, Finn, and
  Levine}]{metaworld}
Yu, T.; Quillen, D.; He, Z.; Julian, R.; Hausman, K.; Finn, C.; and Levine, S.
  2020.
\newblock Meta-world: A benchmark and evaluation for multi-task and meta
  reinforcement learning.
\newblock In \emph{Conference on robot learning}, 1094--1100. PMLR.

\bibitem[{Zahedi et~al.(2022)Zahedi, Sreedharan, Verma, and
  Kambhampati}]{zahedi2022modeling}
Zahedi, Z.; Sreedharan, S.; Verma, M.; and Kambhampati, S. 2022.
\newblock Modeling the Interplay between Human Trust and Monitoring.
\newblock In \emph{2022 17th ACM/IEEE International Conference on Human-Robot
  Interaction (HRI)}, 1119--1123. IEEE.

\bibitem[{Zahedi et~al.(2021)Zahedi, Verma, Sreedharan, and
  Kambhampati}]{zahedi2021trust}
Zahedi, Z.; Verma, M.; Sreedharan, S.; and Kambhampati, S. 2021.
\newblock Trust-aware planning: Modeling trust evolution in longitudinal
  human-robot interaction.
\newblock \emph{arXiv preprint arXiv:2105.01220}.

\end{thebibliography}

\end{document}